\documentclass[twocolumn,10pt,cleanfoot]{asme2e}

\usepackage{graphicx}		% manages graphics in Latex
\usepackage{subfigure}
\usepackage{amsfonts}
\usepackage{amssymb}
\usepackage{amsmath}
\usepackage{color}

\newcommand{\mat}{\mathbf }
\newcommand{\vct}{\boldsymbol }

\newcommand{\tr}{\mathrm{tr}}

\def\tr{\mathrm{tr}}

\confshortname{IDETC2018}
\conffullname{the ASME 2018 International Design Engineering Technical Conferences \& Computers and Information in Engineering Conference}

\confdate{26-29}
\confmonth{August}
\confyear{2018}
\confcity{Quebec City, Quebec}
\confcountry{Canada}

\papernum{DETC2018-85310}

\title{Efficient Load Sampling for Worst-Case Structural Analysis Under Force Location Uncertainty}

\author{Yining Wang, Erva Ulu, Aarti Singh and Levent Burak Kara\thanks{Address all correspondence to this author: \texttt{lkara@cmu.edu}}
    \affiliation{
	Carnegie Mellon University, Pittsburgh PA 15213, USA\\
    }	
}
\begin{document}

\maketitle    

\begin{abstract}
{\it An important task in structural design is to quantify the structural performance of an object under the external forces it may experience during its use. The problem proves to be computationally very challenging as the external forces' contact locations and magnitudes may exhibit significant variations. We present an efficient analysis approach to determine the most critical force contact location in such problems with force location uncertainty. Given an input 3D model and regions on its boundary where arbitrary normal forces may make contact, our algorithm predicts the worst-case force configuration responsible for creating the highest stress within the object. Our approach uses a computationally tractable experimental design method to select number of sample force locations based on geometry only, without inspecting the stress response that requires computationally expensive finite-element analysis. Then, we construct a simple regression model on these samples and corresponding maximum stresses. Combined with a simple ranking based post-processing step, our method provides a practical solution to worst-case structural analysis problem. The results indicate that our approach achieves significant improvements over the existing work and brute force approaches.  We demonstrate that further speed-up can be obtained when small amount of an error tolerance in maximum stress is allowed.
}
\end{abstract}

\section*{INTRODUCTION}

As additive manufacturing technologies become increasingly popular in recent years, optimization in structure design has received much research attention.
%One important sub-problem in lightweight structure optimization is to predict, given a fixed structure and a specific distribution of material mass over the structure,
%the maximum {stress} the structure suffers under worst-case external force applications.
A common approach in formalizing such an optimization problem is to model
the external forces as known and fixed quantities. However, in many real world applications, the
external force's contact locations and magnitudes may change significantly. In order to guarantee
that the resulting structure is robust under all possible force configurations, the maximum
stress experienced within the current shape hypothesis under the worst-case force configuration is to
be computed at each optimization step.
The maximum stress is then compared against a pre-specified tolerance threshold (i.e., material yield strength) and the structure design is then updated accordingly.

Finite-element analysis (FEA) is the standard technique that computes the stress distribution for a given external force configuration and the maximum stress suffered can then be subsequently obtained.
However, FEA is generally computationally expensive, and performing the FEA on every external force configuration possible is out of the question for most structures.
Approximation methods that reduce the total number of FEA runs are mandatory to make the stress prediction and structure optimization problem feasible in practice.

Ulu~\textit{et al.}~\cite{ulu2017lightweight} initiated the research of applying machine learning techniques to the maximum stress prediction problem when there exists uncertainty in the external force locations.
In particular, a small subset of force locations were sub-sampled and the stress responses for forces applied on these locations are computed by FEAs.
Afterwards, a quadratic regression model was built on the sub-sampled data points, which was then used to predict the stress distribution for the other force locations not sampled.
Empirical results show that, with additional post-processing steps, the machine learning based approach gives accurate predictions of the maximum stress using small number of FEAs.

\begin{figure}
\centering
\includegraphics[width=0.96\columnwidth]{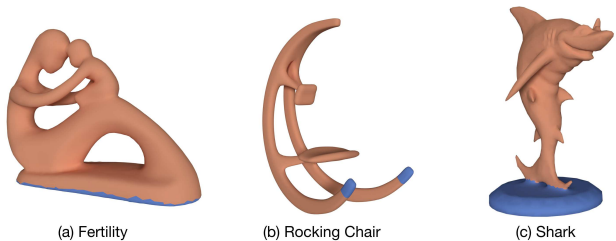}
\caption{Example test structures with complex geometries. Fixed boundary conditions and contact regions are indicated in blue and red, respectively.}
\label{fig:structures}
\end{figure}

Inspired by \cite{ulu2017lightweight}, we propose a new method for predicting the worst-case force configuration for problems in which there is uncertainty in the force locations. Our approach takes as input a 3D shape represented by its boundary surface mesh and a user-specified contact region where external forces may make contact and computes the force configuration that is most likely to result in structural failure. The proposed algorithm pipeline is based on a linear regression model built on the top eigenvectors of the graph Laplacian of the full boundary mesh and a simple ranking post-processing step. We apply a computationally tractable experimental design method to select the training set to reduce the number of FEAs needed without much sacrifice in the prediction accuracy of the worst-case load and the corresponding maximum stress. Our experimental results suggest that the newly proposed algorithm significantly improves over existing work, with approximately $5\times$ fewer FEAs required compared to \cite{ulu2017lightweight}
and up to $100\times$ compared against the brute-force approach.

%In this paper, we improve over the methods in \cite{ulu2017lightweight} to further reduce the number of FEAs needed without much sacrifice in the prediction accuracy of the maximum structure stress.We first propose a simplified algorithm pipeline, which replaces the PCA quadratic regression model in \cite{ulu2017lightweight} by a simple linear regression model built on the top eigenvectors of the graph Laplacian of the full boundary mesh, and substitutes the local search procedure with a simple ranking post-processing step. Furthermore, we apply the computationally tractable experimental design methods developed in \cite{wang2016computational,allenzhu2017near} to select the sub-sampled training set, which improves the naive training set selection algorithm in \cite{ulu2017lightweight} that only maximizes the pairwise geodesic distance between the selected force nodes. Our experimental results suggest that the newly proposed algorithm significantly improves over existing work, with approximately $5\times$ fewer FEAs required compared to \cite{ulu2017lightweight} and up to $100\times$ compared against the brute-force approach.

\section*{RELATED WORK}
Our review focuses on studies that highlight structural analysis and experimental design, with an emphasis on approaches involving structural mechanics.

\subsection*{Structural Analysis}

In predicting structural soundness of an object, stress and deformation analysis using FEA often introduce expensive computational bottlenecks. Commonly, researchers alleviate this issue by using simple elemental structures such as trusses \cite{smith2002creating,rosen2007design,wang2013cost} and beams \cite{zhang2015medial}. For  cases where the structure cannot be represented by these simple elements, Umetani and Schmidt~\cite{umetani2013cross} extend the well-known Euler-Bernoulli model to free-from 3D objects and simplify the problem into 2D cross-sectional analysis.

Specific to problems with force location uncertainty, a common approach in engineering is to use the concept of equivalent uniformly distributed static load to perform simple approximate analysis \cite{choi2002structural}. This concept is commonly encountered in bridge (for traffic load) and building (for wind load) design. However, it is limited to simple geometries, making it unsuitable for our purposes. For complex geometries, Langlois~\textit{et al.}~\cite{langlois2016stochastic} use contact force samples generated by rigid body simulations to predict failure modes of objects in real world use. However, their method is applicable to scenarios where loading is stochastic in nature (such as dropping and collisions) and it is not practical for deterministic scenarios where possible force configurations are known and no failure is tolerated for any of them.

In the context of worst-case structural analysis, Zhou~\textit{et al.}~\cite{zhou2013weak} present a modal analysis based heuristic to static problems in determining the structurally problematic regions that is likely to fail under an arbitrary loading. Building upon this approach, Ulu~\textit{et al.}~\cite{ulu2017lightweight} presents a machine learning technique to find the most critical force configuration efficiently for problems in which there is uncertainty in the force locations. The approach is based on a naive sub-sampling algorithm to select the training set for the machine learning model that maximizes the pairwise geodesic distance between the selected force configurations. Our approach improves over this method by incorporating a computationally tractable experimental design method, resulting in a significant reduction in the number of FEAs needed.

The works of \cite{jalalpour2016efficient,tootkaboni2012topology,zhao2014robust,dunning2011introducing,guo2013robust} consider robust topology optimization methods incorporating uncertainties in material properties, force magnitudes and/or force directions.
On the other hand, in this paper we focus on uncertainties in force locations.
\cite{zhou2013weak} considers a sensitivity based structural analysis approach. Such methods might potentially stuck at local minima,
and could also be slow as many FEAs are required to evaluate the gradients.

\subsection*{Experimental Design}

Experimental design, also known as optimal design, is a classical problem in statistics research \cite{pukelsheim2006optimal,chaloner1995bayesian}.
Given a large pool of potential candidates, a small subset of design points are selected
such that the statistical efficiency by regressing on the selected designs are maximized.
The problem is usually formulated as discrete optimization that is computationally challenging (NP-hard) to solve,
and in practice heuristics such as greedy exchange \cite{fedorov1972theory,miller1994fedorov}
and sampling \cite{wang2016computational} are usually deployed.

Recently, there has been a surge of research in computationally efficient experimental design approaches that enjoy rigorous theoretical guarantees \cite{avron2013faster,horel2014budget,wang2016computational,allenzhu2017near,singh2017personal}.
In this paper, we adopt the methodology developed in \cite{allenzhu2017near}, 
which involves solving a continuous convex optimization problem \cite{joshi2009sensor,wang2016computational} followed by
a greedy rounding algorithm based on graph sparsification techniques \cite{allen2015spectral,silva2016sparse}.
Compared to other methods, the algorithm proposed in \cite{allenzhu2017near}
has the advantages of being applicable to a wide range of optimality criteria and computationally very efficient in practice.

\begin{figure*}
\centering
\includegraphics[width=0.9\textwidth]{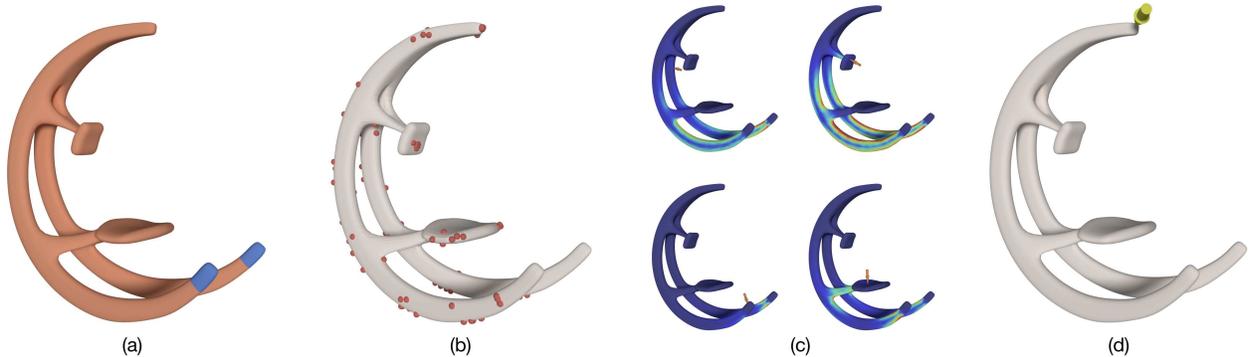}
\caption{Overview of our algorithm. Given a structure (a) with corresponding fixed boundary conditions (blue) and contact regions (red), we use a computationally tractable experimental design method to optimally sample a number of force instants on $\mathcal F$ (b) and perform FEAs to obtain
corresponding maximum stress values (c). We then construct a simple regression model to estimate the largest stresses for the remaining force instants and perform a simple ranking based post-processing step to predict the worst-case force configuration (d). Corresponding maximum stress value constitutes the most critical stress that the object could experience.}
\label{fig:overview}
\end{figure*}

%\section*{PROBLEM FORMULATION AND METHODS}
\section*{PROBLEM FORMULATION}

%In this section, we present the fundamentals and assumptions leading towards the formal description of the problem.

%In this section, we formally formulate the stress prediction problem in lightweight structure design and present a simple pipeline algorithm that produces such a prediction using a small collection of subsampled surface forces. We then introduce the methods developed in \cite{wang2016computational,allenzhu2017near} that subsample the surface elements in a principled way. 

Suppose the input model is represented by a boundary surface mesh $\mathcal S$ and its interior is discretized by a volumetric mesh $\mathcal W$. Our aim is to find the maximum stress generated on $\mathcal W$ for any external force that might make contact within a user-specified contact regions $\mathcal F\subseteq\mathcal S$ (Figure~\ref{fig:structures}). We assume that the object is anchored in space by fixing three or more non-collinear boundary nodes and forces are applied compressively along the surface normal direction.

Under these assumptions, for linear elastic structures, it has been shown in \cite{ulu2017lightweight} that the stress is maximized at some point in $\mathcal W$ when the normal forces are concentrated at a single point rather than being distributed. Therefore, the problem becomes finding the most critical contact point in $\mathcal F$.

%\subsection*{Problem Formulation}

%After triangulation, each structure has $n_S$ \emph{surface regions} $\mathcal S$ and $n_W$ \emph{weak regions} $\mathcal W$.
%Out of the $n_S$ surface regions, there are $n_F<n_S$ \emph{force regions} $\mathcal F\subseteq\mathcal S$ upon which external forces might be applied.
%For example, in the structures depicted in Fig~\ref{fig:structures}, the surface regions on the bottom of the supporting platform are usually excluded from $\mathcal F$
%as no external/user forces are expected there.

%Discretized structure has $n_S$ nodes on the boundary $\mathcal S$ while the entire volumetric mesh $\mathcal W$ contains $n_W$ nodes. Out of the $n_S$ boundary nodes, there are contact regions $\mathcal F\subseteq\mathcal S$ with $n_F < n_S$ nodes upon which external forces might be applied. For example, in the structures depicted in Fig.~\ref{fig:structures}, the objects are fixed from the bottom of the supporting platform (e.g., in the first figure) and corresponding regions are excluded from $\mathcal F$ as no external/user forces are expected there.

Given $\mathcal S,~\mathcal F$ and $\mathcal W$, it is possible to calculate stress distribution over $\mathcal W$ for a force applied at a particular contact point in $\mathcal F$ using FEA, mathematically formulated as
\begin{equation}
\sigma: \mathcal F\times\mathcal W\to\mathbb R_+.
\end{equation}
More specifically, running a single FEA for a force applied on a node $f\in\mathcal F$, one obtains $\sigma(f,\cdot)$, which contains the von Mises stress data $\sigma(f,w)$ for all nodes $w\in\mathcal W$. 
Then, the problem of calculating the maximum stress on $\mathcal W$ incurred by the worst-case external force can be formulated as

%Given the stress distribution $\sigma(f,\cdot)$ for every force node $f\in\mathcal F$, the objective is then to calculate the maximum stress on $\mathcal W$ incurred by the worst-case external force, mathematically defined as
\begin{equation}
\sigma^* := \max_{f\in\mathcal F} \sigma^*(f) := \max_{f\in\mathcal F}~\max_{w\in\mathcal W} \sigma(f,w).
\end{equation}

%\subsection*{The Algorithm Pipeline}
\section*{ALGORITHM}

A naive method to obtain the maximum stress $\sigma^*$ is to compute the stress distribution $\sigma(f,\cdot)$ for every force node $f$ in the contact region $\mathcal F$.
However, as FEAs are computationally heavy and the number of force configurations $n_F$ can be very large, such a brute-force method is infeasible in real-world applications and approximate computations of the maximum stress $\sigma^*$ is mandatory.

\subsection*{Overview}

In this paper, we introduce a computational approach that efficiently computes $\sigma^*$ using a small subset of force nodes in $\mathcal F$.
Our algorithmic pipeline is a great simplification of that developed in \cite{ulu2017lightweight} but yields much more competitive results. 

Figure~\ref{fig:overview} illustrates our approach. From an input 3D shape and prescribed contact regions (Figure~\ref{fig:overview}(a)), our algorithm predicts the force configuration creating the largest stress on the object. We start by sampling a small subset of the contact region $\mathcal F_L\subseteq\mathcal F$ such that $\mathcal F_L$ contains $n_{FL}\ll n_F$ force nodes (Figure~\ref{fig:overview}(b)). After $\mathcal F_L$ is obtained, FEAs are performed on the $n_{FL}$ subsampled force nodes to obtain the stress distribution $\sigma(f,\cdot)$ as well as the maximum stress $\sigma^*(f)$ for all $f\in\mathcal F_L$ (Figure~\ref{fig:overview}(c)). Then, we build a simple linear regression model on these force samples and corresponding maximum stresses and predict the largest stress values for the remaining force nodes. Finally, we perform a simple ranking based post-processing step to make the worst-case force prediction and compute the corresponding maximum stress $\sigma^*$ (Figure~\ref{fig:overview}(d)).

%\paragraph{Subsampling of force nodes}
%The algorithm pipeline starts with sampling a small subset of the contact region $\mathcal F_L\subsete \mathcal F$ such that $\mathcal F_L$ contains $n_{FL}\ll n_F$ force nodes. The subsampling can be accomplished by simple uniform or leverage score sampling \cite{drineas2008relative,spielman2011graph}, or more sophisticated methods such as the $k$-means algorithm \cite{ulu2017lightweight} and the computationally efficient algorithms proposed in \cite{wang2016computational,allenzhu2017near}. After $\mathcal F_L$ is obtained, FEA is performed on the $n_{FL}$ subsampled force nodes to obtain the critical stress map $\sigma(f,\cdot)$  as well as the maximum stress $\sigma^*(f)$ for all $f\in\mathcal F_L$.

\subsection*{Regression Model}
%\paragraph{Linear Laplacian smoothing}

Let $\mat F\in\mathbb R^{n_F\times n_F}$ be the force matrix where each row represents a force configuration. Similar to \cite{ulu2017lightweight}, we distribute the force to a small circular area around the contact point to avoid stress singularities. Therefore, $\mat F(f,f')$ is non-zero if $f'$ 
is inside the circular region centered at $f$ and $\mat F(f,f')=0$ otherwise. Please refer to \cite{ulu2017lightweight} for complete description of the construction of $\mat F$. 

In its original form, dimensionality of the force configuration is very large compared to the number of training samples ($n_{FL}<<n_F$). Moreover, this sparse representation does not carry spatial information, i.e. forces on two spatially very close nodes that create very similar stress distributions can be equally distinct in this representation as two spatially distant forces. To solve these problems, we project the force representation onto a lower dimensional Laplacian basis in which the spatial information is implicitly carried in smaller dimensionality. Let $\mat L\in\mathbb R^{n_F\times p}$ be the top-$p$ eigenvectors of the graph Laplacian of $\mathcal F$, where $p$ is a small integer that is set as $p=15$ in this paper. We refer the readers to \cite{chung1997spectral} for an introduction to the graph Laplacian and its properties. Denote also $\vct\sigma^*=[\sigma^*(f)]_{f\in\mathcal F}\in\mathbb R_+^{n_F}$ as the maximum stress responses for all force configurations in $\mathcal F$. We build the following linear regression model

%Let $\mat F\in\mathbb R^{n_F\times n_F}$ be the \emph{adjacency matrix} on $\mathcal F$, where $\mat F(f,f')$ is the percentage of forces imposed on $f'$ for a small circular force region centered at $f$  if $f,f'$ are neighboring elements or close to each other in $\mathcal F$, and $\mat F(f,f')=0$ otherwise. Sec.~3 in \cite{ulu2017lightweight} gives a complete description of the construction of $\mat F$.
%Let $\mat L\in\mathbb R^{n_F\times p}$ be the top-$p$ eigenvectors of the \emph{graph Laplacian} of $\mathcal F$ or $\mat F$ which captures the spectral properties of the mesh graph constructed on $\mathcal F$, where $p$ is a small integer that is set as $p=15$ in this paper. We refer the readers to the book of \cite{chung1997spectral} for an introduction to the graph Laplacian and its properties. Denote also $\vct\sigma^*=[\sigma^*(f)]_{f\in\mathcal F}\in\mathbb R_+^{n_F}$ as the maximum stress responses for all force configurations in $\mathcal F$.We build the following \emph{linear regression} model: %that models the relatinship between $\vct\sigma^*$ and $\mat F\mat L$:
\begin{equation}
\vct\sigma^* = \overline{\mat F}\mat L\beta_0 + \vct\varepsilon,
\label{eq:model}
\end{equation}
where $\overline{\mat F}$ is the mean-centered $\mat F$ so that each column of $\overline{\mat F}$ sums to 0,
$\beta_0$ is a $p$-dimensional unknown vector that models the linear map, and $\vct\varepsilon\in\mathbb R^{n_F}$ represents modeling noise for each force node.

Let $\vct\sigma^*_L = [\sigma^*(f)]_{f\in\mathcal F_L}$ be the maximum stress responses on the subsampled force nodes in $\mathcal F_L$,
which can be calculated from the results of FEAs carried out on all $f\in\mathcal F_L$.
Let also $[\overline{\mat F}\mat L]_L$ be the corresponding $n_{FL}$ rows in the $n_F\times p$ matrix $\overline{\mat F}\mat L$.
An ordinary least squares (OLS) estimator $\widehat\beta$ can be calculated using $\vct\sigma^*_L$ and $[\overline{\mat F}\mat L]_L$ as follows:
\begin{equation}
\widehat\beta \in \arg\min_{\beta\in\mathbb R^p} \left\|\vct\sigma^*_L - [\overline{\mat F}\mat L]_L\beta\right\|_2^2.
\label{eq:ols}
\end{equation}
Afterwards, a preliminary prediction of maximum stress $\widehat{\vct\sigma}^*$ can be obtained by applying the OLS estimate $\widehat\beta$:
\begin{equation}
\widehat{\vct\sigma}^* = \overline{\mat F}\mat L\widehat\beta.
\label{eq:ols_prediction}
\end{equation}

\subsection*{Force Node Ranking and Final Predictions}

%\paragraph{Force node ranking and final predictions}

Given the maximum stress prediction $\widehat{\vct\sigma}^*$ in Eq.~(\ref{eq:ols_prediction}), it is tempting to
directly report $\max_{f\in\mathcal F}\widehat\sigma^*(f)$ as the final prediction of the maximum stress $\sigma^*$ corresponding to the worst-case force.
However, our empirical results suggest that the absolute estimation error $\|\widehat{\vct\sigma}^*-\vct\sigma^*\|_{\infty}$ is in fact quite large,
which makes the direct predicting approach less desirable.
This phenomenon was also observed in \cite{ulu2017lightweight}, which showed that a quadratic regression model on reduced-dimensional data
has a similar large error in predicting the absolute value of the maximum stress response.

On the other hand, while the absolute estimation error $\|\widehat{\vct\sigma}^*-\vct\sigma^*\|_\infty$ can be large,
we observe that the relative positions of the force nodes which result in maximum stress as predicted by $\widehat{\vct\sigma}^*$ are quite accurate.
In other words, if a force at node $f\in\mathcal F$ is predicted to result in large maximum stress as indicated by $\widehat{\vct\sigma}^*$,
then it is likely to cause large stress in the ground-truth response $\vct\sigma^*$ as well.
This observation motivates us to consider a \emph{ranking} approach for making the final stress predictions.

In particular, for a parameter $k\ll n_F$, we define $\widehat{\mathcal F}_K$ as the top $k$ force nodes in $\mathcal F$ that 
have the largest stress response according to $\widehat{\vct\sigma}^*$.
We then perform FEA on every force node in $\widehat{\mathcal F}_K$, with the maximum stress response denoted as
\begin{equation}
\widehat{\vct\sigma}^*_K = \left[\max_{w\in\mathcal W} \sigma(f,w)\right]_{f\in\widehat{\mathcal F}_K}.
\end{equation}
The predicted maximum stress is then calculated as 
\begin{equation}
\widetilde\sigma^* := \max_{f\in\widehat{\mathcal F}_K}\widehat{\vct\sigma}^*_K(f).
\end{equation}
Compared to the fully $\widehat{\sigma}^*$ based approach, the new predictor $\widehat\sigma_K^*$ is much more accurate as fresh FEAs are run on the predicted ``top'' force nodes;
therefore, as long as the relative force positions resulting in maximum stress as indicated by $\widehat{\vct\sigma}^*$ are sufficiently accurate, the predicted maximum stress $\widetilde\sigma^*$ is very accurate.
For a small $k$, very few number of new FEAs are needed for the final prediction, which is only a small computational overhead incurred on the entire algorithm pipeline.

%\subsection*{Force Node Subsampling by Computationally Tractable Experimental Design}
\subsection*{Computationally Tractable Experimental Design}
\label{subsec:subsampling}

In order to select the contact region subset $\mathcal F_L\subseteq\mathcal F$ that we train the regression model on, we benefit from the experimental design algorithm described in \cite{allenzhu2017near}.
%In this section we give further details on the algorithms proposed in \cite{wang2016computational,allenzhu2017near} that select the contact region subset $\mathcal F_L\subseteq\mathcal F$, corresponding to the first step in our algorithm pipeline.
This approach is based on the (optimal) experimental design question in the statistics literature, and draws tools from the areas of convex optimization and spectral graph sparsification to achieve computational tractability.

%\paragraph{An overview of optimal experimental design}
\paragraph{Optimal experimental design}
%The question of selecting a small number of design points from a large design pool so as to maximize the statistical efficiency on the optimized design is an old topic in statistics, usually termed as \emph{experimental design} or \emph{optimal design} \cite{pukelsheim2006optimal,chaloner1995bayesian}.
%More specifically, 
For the linear regression model $\vct\sigma^*=\mat X\beta_0+\vct\varepsilon$ specified in Eq.~(\ref{eq:model}), 
where $\mat X=\overline{\mat F}\mat L=(x_1,\cdots,x_{n_F})\subseteq\mathbb R^p$, the optimal design problem can be formulated as a combinatorial optimization problem as follows:
\begin{equation}
\min_{s_1,\cdots,s_n} \Phi\left(\sum_{i=1}^n{s_ix_ix_i^\top}\right) \;\;\;\;\;s.t.\;\; s_i\in\{0,1\}, \;\;\sum_{i=1}^n{s_i} \leq n_{FL}.
\label{eq:design}
\end{equation}
Here, the binary variables $\{s_i\}_{i=1}^n$ represent the selected design points (force nodes) in $\mathcal F_L$, and
the objective function $\Phi$ is the optimality criterion that reflects certain aspects of the desired statistical efficiency.
As in our problem where the prediction error $\mat X(\widehat\beta-\beta_0)$ is of primary concern, 
the most relevant criteria are the V- and G-optimality
\begin{itemize}
\item \emph{V-optimality}: $\Phi_V(A) = \frac{1}{n}\tr(XA^{-1}X^\top)$
\item \emph{G-optimality}: $\Phi_G(A) = \max_{1\leq i\leq n}x_i^\top A^{-1}x_i$
\end{itemize}
where $A=\sum_{i=1}^n s_ix_ix_i^\top$ denotes the sample covariance of the selected design points.
The V-optimality $\Phi_V$ measures the variance of the prediction averaged over all design points $x_i$, and the G-optimality $\Phi_G$ measures the maximum prediction variance.
In this paper, we opt for the V-optimality $\Phi_V$ for computational reasons, as $\Phi_V$ is differentiable and hence easier to optimize by first-order optimization methods.

\begin{table}[t]
\footnotesize
\centering
\caption{Statistics of the testing structures}
\vskip 0.1in
\begin{tabular}{ccccc}
\hline
& $n_W$& $n_S$& $n_F$& $\sigma^*$ [MPa]\\
\hline
\textsc{Fertility}& 11221& 4494& 3914& 6.37\\
\textsc{RockingChair}& 15191& 5918& 5348&  37.0\\
\textsc{Shark}& 14152& 5757& 4281& 26.2 \\
\hline
\end{tabular}
\label{tab:statistics}
\end{table}

\paragraph{A continuous relaxation}
As direct optimization of the combinatorial problem in Eq.~(\ref{eq:design}) is difficult, we instead consider a continuous relaxation of it as described in \cite{wang2016computational,allenzhu2017near,joshi2009sensor}:
\begin{equation}
\min_{\pi_1,\cdots,\pi_n} \Phi\left(\sum_{i=1}^n{\pi_ix_ix_i^\top}\right) \;\;\;\;\;s.t.\;\; 0\leq \pi_i\leq 1,\;\; \sum_{i=1}^n\pi_i \leq n_{FL}.
\label{eq:relaxation}
\end{equation}
The only difference here is that the optimization variables are no longer constrained to take integer values. In addition, because the $\Phi_V$ objective is convex, the optimization problem becomes a standard convex continuous optimization problem and can be solved using conventional convex optimization methods. In particular, we use the projected gradient descent method \cite{bubeck2015convex,nesterov2013introductory}, which is observed to converge fast in practice \cite{wang2016computational}. Details of the projected gradient descent algorithm are given in Appendix~A.

\paragraph{Sparsifying the continuous solution}
The optimal solution $\pi^*$ to the continuously relaxed problem in Eq.~(\ref{eq:relaxation}) is in general a dense vector,
and cannot be used directly to obtain a subset $\mathcal F_{L}\subseteq \mathcal F$ with only $n_{FL}$ force nodes.
%A simple fix is to first \emph{normalize} the solution vector $\pi^*$ such that $\sum_{i=1}^n{\pi_i^*}=\sum_{f\in\mathcal F}\pi(f)^* = 1$, and then \emph{sample} $n_{FL}$ force nodes without replacement from $\mathcal F$, where each force node $f\in\mathcal F$ is sampled with probability $\pi(f)^*$ \cite{wang2016computational}. The resulting design subset $\mathcal F_L$ enjoys theoretical guarantees if $n_{FL}$ is not too small compared to $p$. We refer to this method as \textsc{Sampling} in the remainder of this paper.
To address that, we turn the optimal continuous solution $\pi^*$ into a sparse design set as described in \cite{allenzhu2017near}. The algorithm starts with an empty set $\emptyset$ and greedily add elements $f\in\mathcal F$ into the design set $\mathcal F_L$, according to a carefully designed potential function. This greedy algorithm has the advantage of being completely deterministic, thus reducing the uncertainty in sampling methods caused by statistical fluctuation. It has also been shown that the resulting design subset $\mathcal F_L$ enjoys theoretical approximation guarantees. This method will be referred as \textsc{Greedy} in the remainder of the paper and details of the algorithm is given in Appendix~B.

%A more sophisticated approach would be to turn the optimal continuous solution $\pi^*$ into a sparse design set as described in \cite{allenzhu2017near}. The algorithm starts with the empty set $\emptyset$ and greedily add elements $f\in\mathcal F$ into the design set $\mathcal F_L$, according to a carefully designed potential function. This greedy algorithm has the advantage of being completely deterministic, thus reducing the uncertainty in sampling methods caused by statistical fluctuation. It has also been shown that it enjoys better theoretical approximation guarantees compared to the sampling method \cite{allenzhu2017near}. This method will be referred as \textsc{Greedy} in the remainder of the paper and we attach the details of the algorithm in Appendix~B.

\begin{table*}[t]
\footnotesize
\centering
\caption{Results for the \textsc{Fertility} model. Randomized algorithms (\textsc{Uniform}, \textsc{Levscore} and \textsc{Sampling}) are run for 10 independent trials and the median performance is reported. Best performing settings are indicated in bold.
Total FEAs equals $n_{FL}$ plus numbers in the table.}
\vskip 0.1in
\begin{tabular}{llcccccccc}
\hline
& \multicolumn{1}{r}{$n_{FL}=$}& 25& 50& 100& 150& 200& 250& 300&  Total FEAs\\
\hline
$\delta=0$& 
\textsc{Uniform}& 316.5& 149& 78.5& 37.5& 98.5& 42.5& 39& 178.5 ($n_{FL}=100$)\\
& \textsc{Levscore}& 252.5& 54.5& 73.5& 68.5& 42.5& 31& 13.5& 104.5 ($n_{FL}=50$)\\
& \textsc{K-means}& 237& {\bf 25}& 61& 82& 57& {\bf 17}& {\bf 16}& 75 ($n_{FL}=50$)\\
& \textsc{Sampling}& 210.5& 148.5& 51& 30& 35.5& 34& 26.5& 151 ($n_{FL}=100$)\\
& \textsc{Greedy}& {\bf 12}& 26& {\bf 13}& {\bf 7}& {\bf 11}& 25& {\bf 33}& {\bf 37 ($n_{FL}=25$)}\\
\hline
$\delta=0.05$& 
\textsc{Uniform}& 285& 80.5& 52& 10& 63& 10& 10& 130.5 ($n_{FL}=50$)\\
&\textsc{Levscore}& 175& 26.5& 55.5& 59& 17& 10& 7& 76.5 ($n_{FL}=50$)\\
&\textsc{K-means}& 144& {\bf 2}& 19& 22& 14& {\bf 2}& {\bf 2}& 52 ($n_{FL}=50$)\\
&\textsc{Sampling}& 202& 113& 10& {\bf 7}& 11& 8& 6& 110 ($n_{FL}=100$)\\
&\textsc{Greedy}& {\bf 4}& 3& {\bf 4}& {\bf 7}& {\bf 5}& {\bf 2}& 6& {\bf 29 ($n_{FL}=25$)}\\
\hline
$\delta = 0.1$& 
\textsc{Uniform}& 87.5& 37.5& 13& 7& 15.5& 7& 6.5& 87.5 ($n_{FL}=50$)\\
&\textsc{Levscore}& 59& 13& 15& 14& 13& 8& 6& 63 ($n_{FL}=50$)\\
&\textsc{K-means}& 46& {\bf 1}& 7& 20& 8& {\bf 1}& {\bf 1}& 51 ($n_{FL}=50$)\\
&\textsc{Sampling}& 88& 25& 7.5& 6& 7.5& 6.5& 4& 75 ($n_{FL}=50$)\\
&\textsc{Greedy}& {\bf 4}& {\bf 1}& {\bf 1}& {\bf 4}& {\bf 1}& {\bf 1}& 4& {\bf 29 ($n_{FL}=25$)}\\
\hline
\end{tabular}
\label{tab:fertility}
\end{table*}

\begin{table*}[!h]
\footnotesize
\centering
\caption{Results for the \textsc{RockingChair} model. Randomized algorithms (\textsc{Uniform}, \textsc{Levscore} and \textsc{Sampling}) are run for 10 independent trials and the median performance is reported. Best performing settings are indicated in bold.
Total FEAs equals $n_{FL}$ plus numbers in the table.}
\vskip 0.1in
\begin{tabular}{llcccccccc}
\hline
& \multicolumn{1}{r}{$n_{FL}=$}& 25& 50& 100& 150& 200& 250& 300&  Total FEAs\\
\hline
$\delta=0$& 
\textsc{Uniform}& 716& 857& 385.5& 42& 135.5& 269.5& {\bf 36}& 192 ($n_{FL}=150$)\\
& \textsc{Levscore}& 764.5& 208.5& {\bf 36}& {\bf 36}& {\bf 36}& {\bf 36}& {\bf 36}& 136 ($n_{FL}=100$)\\
& \textsc{K-means}& 4013& 4400& 4573& 4301& 4320& 4620& 4757& 4038 ($n_{FL}=25$)\\
& \textsc{Sampling}& 672.5& 282& 38.5& 38& 38& {\bf 36}& {\bf 36}& 138.5 ($n_{FL}=100$)\\
& \textsc{Greedy}& {\bf 36}& {\bf 35}& 208& {\bf 35}& {\bf 36}& {\bf 36}& {\bf 36}& {\bf 61 ($n_{FL}=25$)}\\
\hline
$\delta=0.05$& 
\textsc{Uniform}& 404& 466& 201.5& 20& 88& 93.5& 18& 170 ($n_{FL}=150$)\\
&\textsc{Levscore}& 444& 192.5& 20& {\bf 18.5}& {\bf 18}& {\bf 18}& {\bf 18}& 120 ($n_{FL}=100$)\\
&\textsc{K-means}& 285& 466& {\bf 14}& 24& 26& 161& 195& 114 ($n_{FL}=100$)\\
&\textsc{Sampling}& 540& 268& 21.5& 20.5& 20.5& 20& 20& 121.5 ($n_{FL}=100$)\\
&\textsc{Greedy}& {\bf 20}& {\bf 19}& 200& 20& 20& 20& 20& {\bf 45 ($n_{FL}=25$)}\\
\hline
$\delta = 0.1$& 
\textsc{Uniform}& 399& 384.5& 135.5& 15.5& 60.5& 75.5& 14& 165.5 ($n_{FL}=150$)\\
&\textsc{Levscore}& 437& 145& 14& 14& {\bf 14}& {\bf 14}& {\bf 14}& 114 ($n_{FL}=100$)\\
&\textsc{K-means}& 258& 395& {\bf 5}& {\bf 13}& 16& 140& 184& 105 ($n_{FL}=100$)\\
&\textsc{Sampling}& 535& 237& 16& 16& 16& 16& 16& 116 ($n_{FL}=100$)\\
&\textsc{Greedy}& {\bf 14}& {\bf 14}& 175& 16& 16& {\bf 14}& {\bf 14}& {\bf 39 ($n_{FL}=25$)}\\
\hline
\end{tabular}
\label{tab:rockingchair}
\end{table*}

\begin{table*}[t]
\footnotesize
\centering
\caption{Results for the \textsc{Shark} model. Randomized algorithms (\textsc{Uniform}, \textsc{Levscore} and \textsc{Sampling}) are run for 10 independent trials and the median performance is reported. Best performing settings are indicated in bold.
Total FEAs equals $n_{FL}$ plus numbers in the table.}
\vskip 0.1in
\begin{tabular}{llcccccccc}
\hline
& \multicolumn{1}{r}{$n_{FL}=$}& 25& 50& 100& 150& 200& 250& 300&  Total FEAs\\
\hline
$\delta=0$& 
\textsc{Uniform}& 585& 384& 141.5& 208.5& 20&\bf 9& 9.5& 220 ($n_{FL}=200$)\\
& \textsc{Levscore}& 478.5& \bf 9& \bf 9&\bf 9&\bf 9& \bf9& \bf9& 59 ($n_{FL}=50$)\\
& \textsc{K-means}& 133& 102&\bf 9& \bf9&\bf 9&\bf 9&\bf 9& 109 ($n_{FL}=100$)\\
& \textsc{Sampling}&963.5& 87&\bf 9& \bf9&\bf 9&\bf 9&\bf 9& 109 ($n_{FL}=100$)\\
& \textsc{Greedy}& \bf9& 171&\bf 9& \bf9& \bf9& \bf9&\bf 9& \bf34 ($n_{FL}=25$)\\
\hline
$\delta=0.01$& 
\textsc{Uniform}& 568.5& 341& 131.5& 156& 15& \bf4& 4.5& 215 ($n_{FL}=200$)\\
&\textsc{Levscore}& 416& \bf4& \bf4& \bf4&\bf 4& \bf4& \bf4& 54 ($n_{FL}=50$)\\
&\textsc{K-means}& 129& 84& \bf4& \bf4& \bf4&\bf 4& \bf4& 104 ($n_{FL}=100$)\\
&\textsc{Sampling}& 872.5& 69& \bf4& \bf4&\bf 4& \bf4&\bf 4& 104 ($n_{FL}=100$)\\
&\textsc{Greedy}& \bf4& 115& \bf4&\bf 4&\bf 4&\bf 4& \bf4& \bf 29 ($n_{FL}=25$)\\
\hline
$\delta = 0.05$& 
\textsc{Uniform}& 323& 172.5& 52& 75& 10&\bf 4& 4.5& 151 ($n_{FL}=100$)\\
&\textsc{Levscore}& 225& \bf4&\bf 4& \bf4& \bf4&\bf 4&\bf 4& 54 ($n_{FL}=50$)\\
&\textsc{K-means}& 129& 80&\bf 4&\bf 4&\bf 4&\bf 4&\bf 4& 104 ($n_{FL}=100$)\\
&\textsc{Sampling}& 391.5& 52.5&\bf 4&\bf 4&\bf 4&\bf 4&\bf 4& 102.5 ($n_{FL}=50$)\\
&\textsc{Greedy}& \bf 4& 115&\bf 4&\bf 4&\bf 4&\bf 4&\bf 4&\bf 29 ($n_{FL}=25$)\\
\hline
\end{tabular}
\label{tab:shark}
\end{table*}

\begin{figure*}[!h]
\centering
\includegraphics[width=0.7\textwidth]{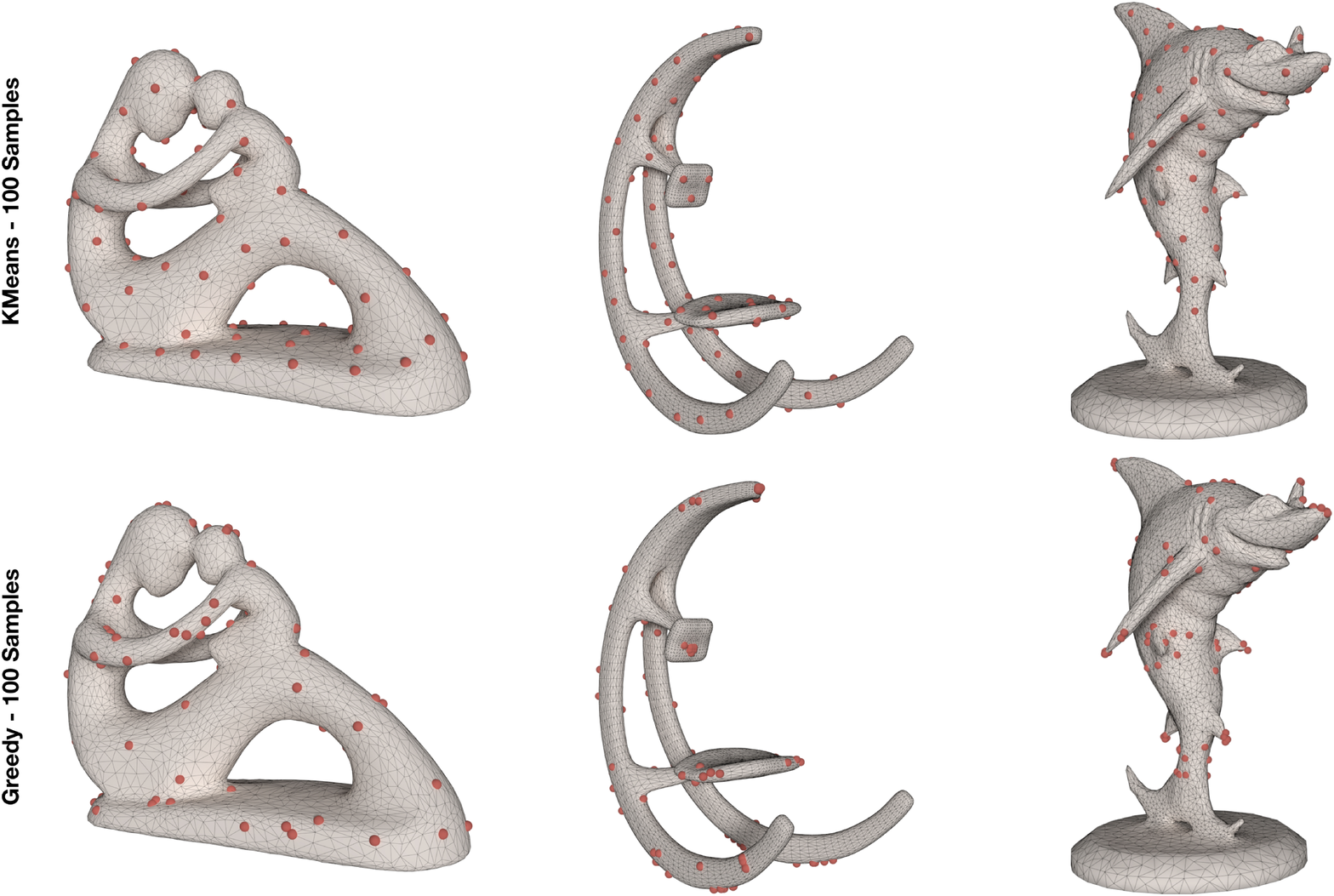}
\caption{Sampled force nodes ($\mathcal F_L$) using the  \textsc{K-means} algorithm (top row) versus the  \textsc{Greedy} algorithm (bottom row)
for $n_{FL}=100$.}
\label{fig:k100}
\end{figure*}

\begin{figure*}[t]
\centering
\includegraphics[width=0.7\textwidth]{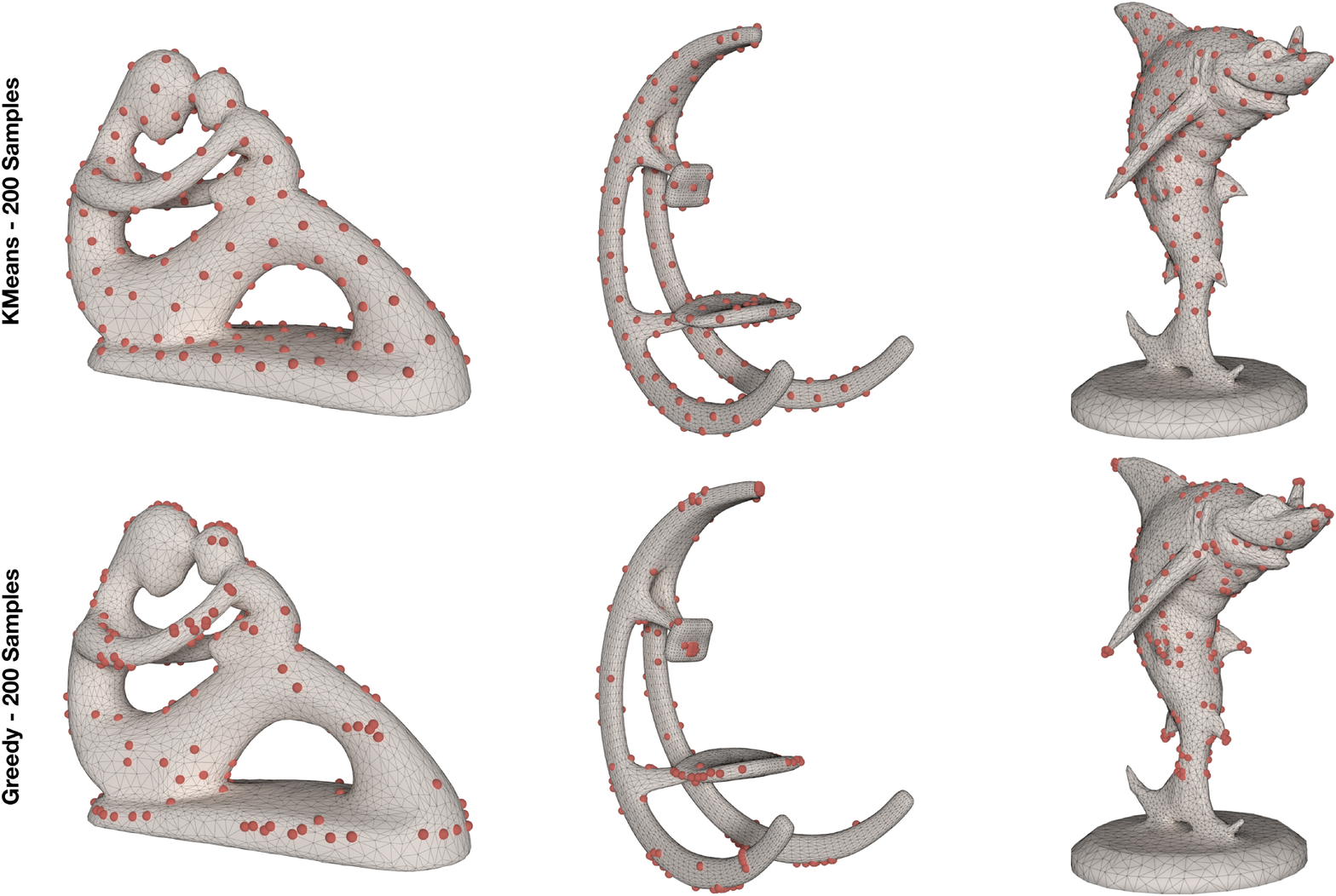}
\caption{Sampled force nodes ($\mathcal F_L$) using the  \textsc{K-means} algorithm (top row) versus the \textsc{Greedy} algorithm (bottom row) for $n_{FL}=200$.}
\label{fig:k200}
\end{figure*}

\section*{EXPERIMENTS}

%\subsection*{Data Descriptions and Methods}
%\label{subsec:data}

We evaluate the performance of our algorithm on three test structures (\textsc{Fertility}, \textsc{RockingChair} and \textsc{Shark}) illustrated in Fig.~\ref{fig:structures}.
%We use three test structures (\textsc{Fertility}, \textsc{RockingChair} and \textsc{Shark}) to evaluate the performance of our proposed algorithm pipeline.
Table~\ref{tab:statistics} gives a description of statistics of these test structures, including number of nodes in $\mathcal W$, $\mathcal S$ and $\mathcal F$ as $n_W$, $n_S$ and $n_F$ and the maximum stress $\sigma^*$.

In our experiments, we consider 5 different methods to sample the force nodes $\mathcal F_L$, which is arguably the most important step in our algorithmic framework. We compare the $\textsc{Greedy}$ approach described previously in this paper with relatively simple baseline methods, $\textsc{Uniform}$ and $\textsc{Levscore}$, as well as the previous work, $\textsc{K-means}$ \cite{ulu2017lightweight} and $\textsc{Sampling}$ \cite{wang2016computational} :

%Within our algorithm pipeline, we consider 5 different methods that subsample the force nodes $\mathcal F_L$, which is arguably the most important step in our algorithmic framework. The first two methods, $\textsc{Uniform}$ and $\textsc{Levscore}$, are the relatively simple uniform and leverage score sampling methods that serve as baselines. The third method $\textsc{K-means}$ produces force samples that are as uniform as possible on the structure surface, and was the subsampling method used in previous work \cite{ulu2017lightweight}. The last two methods correspond to the \textsc{Sampling} and \textsc{Greedy} algorithms described in Sec.~\ref{subsec:subsampling} in this paper.

\begin{enumerate}

\item \textsc{Uniform}: The sample set $\mathcal F_L$ is obtained by sampling without replacement each force node $f\in\mathcal F$ uniformly at random, until $n_{FL}$ samples are obtained.

\item \textsc{Levscore}: The sample set $\mathcal F_L$ is obtained by sampling without replacement each force node $f\in\mathcal F$ with probability proportional to its leverage score,
defined as $x(f)^\top (\mat X^\top\mat X)^{-1}x(f)$, until $n_{FL}$ samples are obtained.

\item \textsc{K-means}: The sample set $\mathcal F_L$ consists of $n_{FL}$ force nodes in $\mathcal F$ such that the geodesic distance between the closest force nodes in $\mathcal F_L$ is maximized.
As the problem itself is NP-hard, the K-means (Lloyd's) algorithm is employed to get an approximate solution.

\item\textsc{Sampling}: The sample set $\mathcal F_L$ is obtained by sampling without replacement each force node $f\in\mathcal F$ with probability $\pi(f)^*$ until $n_{FL}$ samples are obtained, where $\pi^*$ is the optimal solution to Eq.~(\ref{eq:relaxation}).

%\item\textsc{Greedy}: The sample set $\mathcal F_L$ is obtained by the greedy selection algorithm described in \cite{allenzhu2017near}, which is based on the optimal solution $\pi^*$ to Eq.~(\ref{eq:relaxation}).

\end{enumerate}

\subsection*{Evaluation Measures}

%Suppose $\widetilde\sigma_K^*$ is the predicted maximum stress, which by definition is always less than or equal to the ground truth $\sigma^*$.
The predicted maximum stress $\widetilde\sigma^*$ is by definition less than or equal to the ground truth $\sigma^*$. 
The performance of an algorithm is evaluated by the smallest integer $k$ required such that $\sigma^*\leq (1+\delta) \widetilde\sigma^*$, where $\delta\geq 0$ is a pre-specified
error tolerance parameter.
The setting of $\delta=0$ asks for exact identification of the maximum stress caused by the worst-case external force application, while $\delta>0$ allows for a small error margin in the prediction of $\widetilde\sigma^*$.

%\subsection*{Results}
\section*{RESULTS AND DISCUSSION}

We report the smallest $k$ recovered for $\widetilde\sigma^*$ to be lower bounded by $\sigma^*/(1+\delta)$ in Tables~\ref{tab:fertility}, \ref{tab:rockingchair} and \ref{tab:shark}
for the three test structures.
We report the performance for $\delta\in\{0,0.01, 0.05, 0.1\}$ settings, and the sizes of the sub-sampled training set $n_{FL}$ ranging from 25 to 300.
%with larger $n_{FL}$ generally resulting in more accurate prediction (and hence smaller $K$) but computationally more expensive.
The total number of FEAs needed by an algorithm is computed as the sum of $n_{FL}$ and $k$.
For non-deterministic algorithms (\textsc{Uniform}, \textsc{Levscore} and \textsc{Sampling}), we perform 10 repetitions and report the median performance.

Tables \ref{tab:fertility}, \ref{tab:rockingchair} and \ref{tab:shark} suggest that both the \textsc{K-means} and the \textsc{Greedy} algorithms outperform their competitors for most parameter settings.
One important reason for the overall good performance of \textsc{K-means} and \textsc{Greedy} is their deterministic nature, which avoids poor designs due to statistical perturbations in the other randomized algorithms.
Furthermore, the \textsc{Greedy} algorithm remains accurate and robust even when $n_{FL}$ is very small (e.g., $n_{FL}=25$). For such a small $n_{FL}$ setting, the other methods require large $k$ values to compensate for the prediction error.
Therefore, the \textsc{Greedy} algorithm can produce an accurate prediction of the overall maximum stress using much fewer number of total FEAs in both the first and the last stages 
of our algorithm pipeline, as shown by the rightmost columns in the tables. Our approach uses no more than $65$ FEAs to successfully recover the maximum stress caused by worst-case external forces. In addition, when a $5\%$ to $10\%$ error tolerance is used, the number of FEAs can be further reduced to less than $40$. This is close to a $100\times$ speed-up compared to the brute-force approach that performs FEA on the entire surface mesh. It also achieves a $5\times$ speed-up over the existing work \cite{ulu2017lightweight} and is simpler to implement.

%\section*{DISCUSSION}

In Fig.~\ref{fig:k100} and Fig.~\ref{fig:k200}, we plot the sub-sampled force nodes (i.e., $\mathcal F_L$) of the \textsc{Greedy} algorithm for $n_{FL}=100$ and $n_{FL}=200$, respectively. We provide the samples obtained by the \textsc{K-means} algorithm in \cite{ulu2017lightweight} for comparison.
The difference in the sampling patterns between \textsc{Greedy} and \textsc{K-means} are quite obvious from the figures. We explain the differences for the three structures separately:
\begin{itemize}
\setlength\itemsep{1em}
\item \textsc{Fertility}: The \textsc{K-means} algorithm emphasizes the pairwise geodesic distance between sample points and thus places samples in a uniform fashion on the bodies, necks and heads of the structure.
On the other hand, the \textsc{Greedy} algorithm places more samples on the arms connecting the mother and the child, which are the most fragile parts of the structure.
By placing more samples on these parts the learned linear model is more accurate in predicting the maximum stress, and therefore fewer FEAs are required to achieve a certain error tolerance level
in $\widetilde\sigma^*$.

\item \textsc{RockingChair}: The \textsc{Greedy} algorithm produces more samples on the top end of the body, the surface area of the smaller back support and the edges of the larger seat compared to the equally distanced \textsc{K-means} design.
External forces applied onto these parts of this structure are likely to cause increased stress, and therefore more samples placed around this region can greatly improve the regression model
built for the maximum stress.

\item\textsc{Shark}: As reported in Table~\ref{tab:shark}, most of the sampling methods can predict the maximum stress using very small number of FEAs. However, if we focus on the sample points on the fins of the shark there are some noticeable differences between \textsc{K-means} and \textsc{Greedy}. While the \textsc{Greedy} algorithm places more points around the tips, samples produced by the \textsc{K-means} algorithm are relatively uniformly distributed on the surfaces. This subtle difference makes \textsc{Greedy} algorithm more robust in prediction accuracy for small $n_{FL}$ values.
\end{itemize}

Despite the significant reduction in FEA time, one important limitation of the proposed algorithm is the lack of stopping criteria. In particular, the performance parameter $k$ can only be evaluated once the ground-truth maximum stress $\sigma^*$ is known. On the other hand, performance control in problems involving structural mechanics is of vital importance because designs with insufficient stress tolerance may actually fail in reality, leading to devastating consequences. In our examples, we empirically determined that $k = 40$ is sufficient for most of the $n_{FL}$ and $\delta$ settings. Determining the optimum value in a more principled way remains as an open problem and an immediate future work.

\section*{CONCLUSION}
We present an efficient analysis approach for 3D objects under force location uncertainty. The proposed worst-case analysis approach efficiently determines the force contact point creating the highest stress in the structure. Driven by a computationally tractable experimental design method, we show that the relationship between the force configurations and resulting largest stress can be captured using only small number of FEA evaluations. We evaluate the performance of our algorithm on a set of arbitrarily complex 3D models. The results indicate that significant improvements over a brute force approach and an existing work can be achieved. 

\bibliographystyle{asmems4}

\begin{acknowledgment}
We are grateful to the designers whose 3D models we used: Aim@Shape for the fertility, Qingnan Zhou for the rocking chair and Lu et al. for the shark. This work is partly funded by NCDMM America Makes Project \#4058.
\end{acknowledgment}

\bibliography{mmwuols}

%%%%%%%%%%%%%%%%%%%%%%%%%%%%%%%%%%%%%%%%%%%%%%%%%%%%%%%%%%%%%%%%%%%%%%
\appendix

\section*{Appendix A: The Projected Gradient Descent Algorithm}
\label{AppA}

Using the $V$-optimality $\Phi_V(A)=\tr(XA^{-1}X^\top)$, the partial derivative of $\Phi_V$ with respect to $\pi_i$ can be calculated as
\begin{equation}
\frac{\partial\Phi_V}{\partial\pi_i} = -\frac{1}{n}x_i^\top \boldsymbol{\Sigma}^{-1}\mat X^\top\mat X\boldsymbol{\Sigma}^{-1}x_i,
\label{eq:grad}
\end{equation}
where $\boldsymbol \Sigma = \sum_{j=1}^n{\pi_ix_ix_i^\top}$.
Let also $\Pi:=\{\pi\in\mathbb R^n: 0\leq\pi_i\leq 1, \sum_{i=1}^n{\pi_i}\leq n_{FL}\}$ be the feasible set.
The projected gradient descent algorithm can then be formulated as following:
\begin{enumerate}
\item \textbf{Input}: $\mat X\in\mathbb R^{n_F\times p}$, feasibility set $\Pi$, algorithm parameters $\alpha\in(0,1/2]$, $\beta\in(0,1)$;
\item \textbf{Initialization}: $\pi^{(0)}=(n_{FL}/n_F, \cdots, n_{FL}/n_F)$, $t=0$;
\item \textbf{While} stopping criteria are not met \textbf{do the following}:
\begin{enumerate}
\item Compute the gradient $g_t=\nabla_\pi\Phi(\pi^{(t)})$ using Eq.~(\ref{eq:grad});
\item Find the smallest integer $s\geq 0$ such that $\Phi(\pi)-\Phi(\pi^{(t)})\leq \alpha g_t^\top(\pi-\pi^{(t)})$, where $\pi=\mathcal P_\Pi(\pi^{(t)}-\beta^sg_t)$;
\item Update: $\pi^{(t+1)}=\mathcal P_\Pi(\pi^{(t)}-\beta^sg_t)$, $t\gets t+1$. 
\end{enumerate}
\end{enumerate}
Here in Steps 3(b) and 3(c), the $\mathcal P_\Pi(\cdot)$ is the projection operator onto the (convex) constrain set $\Pi$ in Euclidean norm.
More specifically, $\mathcal P_\Pi(\pi) := \arg\min_{z\in\Pi}\|\pi-z\|_2$.
Such projection can be efficiently computed in almost linear time up to high accuracy \cite{wang2016computational,duchi2008efficient,condat2015fast,su2012efficient}.
Also, the step 3(b) corresponds to the Amijo's rule (also known as backtracking line search) that automatically selects step sizes,
a popular and efficient method for step size selection in full gradient descent methods.

\section*{Appendix B: The Greedy Algorithm}

The \textsc{Greedy} algorithm was proposed in \cite{allenzhu2017near} as a principled method to sparsify the continuous optimization solution $\pi^*$.
The algorithm makes uses of a carefully designed potential function for $i\in[n_F]$ and $\Lambda\subseteq[n_F]$: 
\begin{equation}
\begin{aligned}
&\psi(i;\Lambda) := \frac{x_i^\top\mat B(\Lambda) x_i}{1+\alpha x_i^\top \mat B(\Lambda)^{1/2}x_i} \\
\text{where}~~&\mat B(\Lambda) = \left(c\mat I + \sum_{j\in\Lambda}x_jx_j^\top\right)^{-2}, ~~\tr(\mat B(\Lambda))=1.
\end{aligned}
\label{eq:potential}
\end{equation}
Here, $\alpha>0$ and $c\in\mathbb R$ is the unique real number such that $\tr(\mat B(\Lambda))=1$.
The exact values of $c$ can be computed efficiently using a binary search procedure, as shown in \cite{allenzhu2017near}.
The potential function is inspired by a regret minimization interpretation of the least singular values of sum of rank-1 matrices.
Interested readers should refer to \cite{allenzhu2017near,silva2016sparse,allen2015spectral} for more details.

Based on the potential function in Eq.~(\ref{eq:potential}), the \textsc{Greedy} algorithm starts with an empty set and add force nodes one by one in a greedy manner,
until there are $n_{FL}$ elements in $\mathcal F_L$.
\begin{enumerate}
\item \textbf{Input}: $\mat X\in\mathbb R^{n_F\times p}$, budget $n_{FL}$, optimal solution $\pi^*$, algorithm parameter $\alpha>0$;
\item \textbf{Whitening}: $\mat X\gets \mat X(\mat X \boldsymbol{\Sigma}_*\mat X^\top)^{-1/2}$, where $\boldsymbol{\Sigma}_* = \sum_{i=1}^n{\pi_i^*x_ix_i^\top}$;
\item\textbf{Initialization}: $\Lambda_0=\emptyset$, $\mathcal F_L=\emptyset$;
\item \textbf{For} $t=1$ to $n_{FL}$ \textbf{do the following}:
\begin{enumerate}
\item Compute $\psi(i;\Lambda_{t-1})$ for all $i\notin\Lambda_{t-1}$ and select $i_t := \arg\max_{i\notin\Lambda_{t-1}}\psi(i;\Lambda_{t-1})$;
\item Update: $\Lambda_t = \Lambda_{t-1}\cup\{i_t\}$, $\mathcal F_L\gets \mathcal F_L \cup\{i_t\}$.
\end{enumerate}
\end{enumerate}

\end{document}